\documentclass{article}

\usepackage{microtype}
\usepackage{graphicx}
\usepackage{subfigure}
\usepackage{booktabs} 
\usepackage{colortbl}  

\usepackage{hyperref}

\usepackage[accepted]{icml2025}

\usepackage{amsmath}
\usepackage{amssymb}
\usepackage{mathtools}
\usepackage{amsthm}
\usepackage{xcolor}
\usepackage{subcaption}

\usepackage{booktabs} 
\usepackage{multirow} 

\usepackage[capitalize,noabbrev]{cleveref}

\theoremstyle{plain}

\theoremstyle{definition}

\theoremstyle{remark}

\usepackage[textsize=tiny]{todonotes}

\icmltitlerunning{FedEAT: A Robustness Optimization Framework for Federated LLMs}

\begin{document}

\twocolumn[
\icmltitle{FedEAT: A Robustness Optimization Framework for Federated LLMs}




\begin{icmlauthorlist}
\icmlauthor{Yahao Pang}{yyy,comp}
\icmlauthor{Xingyuan Wu}{comp}
\icmlauthor{Wei Chen}{comp}
\icmlauthor{Hai Jin}{yyy,comp}
\icmlauthor{Xiaojin Zhang}{yyy,comp}
\end{icmlauthorlist}

\icmlaffiliation{yyy}{National Engineering Research Center for Big Data Technology and System, Services Computing Technology and System Lab, Cluster and Grid Computing Lab, School of Computer Science and Technology, Huazhong University of Science and Technology, Wuhan, 430074, China}
\icmlaffiliation{comp}{Huazhong University of Science and Technology, Wuhan, China}

\icmlcorrespondingauthor{Xiaojin Zhang}{xiaojinzhang@hust.edu.cn}

\icmlkeywords{Machine Learning, ICML}

\vskip 0.3in
]



\printAffiliationsAndNotice{}  

\begin{abstract}
Significant advancements have been made by Large Language Models (LLMs) in the domains of natural language understanding and automated content creation. However, they still face persistent problems, including substantial computational costs and inadequate availability of training data. The combination of Federated Learning (FL) and LLMs (federated LLMs) offers a solution by leveraging distributed data while protecting privacy, which positions it as an ideal choice for sensitive domains. 
However, Federated LLMs still suffer from robustness challenges, including data heterogeneity, malicious clients, and adversarial attacks, which greatly hinder their applications. We first introduce the robustness problems in federated LLMs, to address these challenges, we propose FedEAT (Federated Embedding space Adversarial Training), a novel framework that applies adversarial training in the embedding space of client LLM and employs a robust aggregation approach, specifically geometric median aggregation, to enhance the robustness of Federated LLMs. Our experiments demonstrate that FedEAT effectively improves the robustness of Federated LLMs with minimal performance loss. 
\end{abstract}

\section{Introduction}
Large Language Models have achieved substantial breakthroughs and are currently extensively utilized across multiple domains, including natural language comprehension, code editing, and text generation \cite{nam2024using,tang2024science,yao2024survey}. However, these models 
face challenges related to high computational demands and insufficient training data \cite{YAO2024100211,minaee2024large}. Federated large-language models (federated LLMs) effectively use decentralized data and distributed computing resources to build powerful models without violating privacy regulations, making them particularly useful in privacy-sensitive fields\cite{chen2024integrationlargelanguagemodels,10.1007/978-981-97-5569-1_17,sani2024futurelargelanguagemodel}.

Previous research \cite{xhonneux2024efficientadversarialtrainingllms,mazeika2024harmbenchstandardizedevaluationframework,lyu2022privacy,10571602} has shown that both FL and LLM are affected by robustness, which denotes a system's capability to sustain consistent functionality and output quality despite encountering data variations, environmental disturbances, or potential adversarial manipulations. In FL, data heterogeneity and malicious clients can substantially impair the overall effectiveness of the global model, potentially leading to erroneous output. Meanwhile, LLMs can generate misleading or even harmful content when faced with adversarial examples or out-of-distribution data \cite{liu2023trustworthy}. If attackers exploit these vulnerabilities, they could not only result in undesirable model outputs but also lead to privacy breaches, security risks, or decision-making errors. Therefore, robustness has become a critical performance metric that must be addressed in the practical application of FL and LLM.

Federated LLMs face the robustness challenges of both FL and LLMs simultaneously. On the one hand, federated LLMs must ensure global effectiveness and robustness in the presence of data heterogeneity and malicious clients. On the other hand, they must address the risks posed by adversarial examples and out-of-distribution data. In a distributed environment, attackers might launch attacks from multiple clients simultaneously, further exacerbating robustness challenges. Additionally, federated LLMs must enhance robustness while controlling communication and computational overhead to ensure practical usability in distributed settings. Efficiently addressing these dual robustness challenges is one of the core challenges facing federated LLMs.

Adversarial training is a proven method for enhancing model robustness and has been well-established in Machine Learning \cite{bai2021recent,zhao2022adversarial} and FL \cite{zizzo2020fat,zhang2023delving}. In the realm of LLMs, efforts have begun to take advantage of adversarial training to improve their robustness against adversarial attacks \cite{mazeika2024harmbenchstandardizedevaluationframework}. However, the exploration of integrating adversarial training into federated LLMs remains relatively limited. In federated LLMs, the challenges of distributed data heterogeneity, communication constraints, and the sheer scale of the models present new obstacles to the design and implementation of adversarial training. There is an urgent need to develop efficient adversarial training methods tailored to federated LLMs to enhance its robustness.

  
The key contributions of this work include: 
\begin{itemize}  
    \item We systematically review the key robustness challenges faced by federated LLMs, including robustness challenges inherited from both FL and LLMs, as well as new challenges arising from their integration.  
    \item We propose the FedEAT algorithm, which enhances the robustness of federated large models by applying adversarial training in the embedding space combined with a federated aggregation method using geometric median.  
    \item Extensive experiments validate the effectiveness of the FedEAT algorithm in improving robustness, particularly against adversarial attacks, while minimizing performance degradation.  
\end{itemize}

\section{Related Works}\label{sec2}
This section systematically explores relevant research work in the areas of Federated Learning (FL), Large Language Models (LLMs), Federated LLMs, and adversarial training.

\subsection{Federated Learning}



FL faces significant robustness challenges, primarily related to the following challenges: One major problem is data heterogeneity, where clients may possess data distributions that vary significantly, making it difficult to achieve a global optimal model \cite{lyu2022privacy}. Additionally, malicious clients \cite{li2020learning} can pose a threat to the global model’s performance by uploading poisoned model updates. Adversarial attacks \cite{pmlr-v97-bhagoji19a} also represent a serious challenge, as attackers can manipulate local updates to introduce vulnerabilities into the model. To address these challenges, several methods have been proposed, such as robust aggregation techniques \cite{pillutla2022robust}, adversarial training \cite{zizzo2020fat}, and anomaly detection \cite{mothukuri2021federated}. However, enhancing the robustness of FL models remains an active area of research.

\subsection{Large Language Models}

LLMs face challenges in terms of robustness, particularly when dealing with out-of-distribution data or adversarial input \cite{liu2023trustworthy,minaee2024largelanguagemodelssurvey}. Previous research has shown that LLMs are highly sensitive to minor perturbations in input data, which may result in a significant drop in performance. Methods to enhance the robustness of LLMs often involve adversarial training, data augmentation, and regularization. Despite these advancements, LLMs remain vulnerable to various adversarial attacks, making robustness a critical area of concern.

\subsection{Federated Large Language Models}
\begin{figure}[htbp]  
    \centering
    \includegraphics[width=0.4\textwidth]{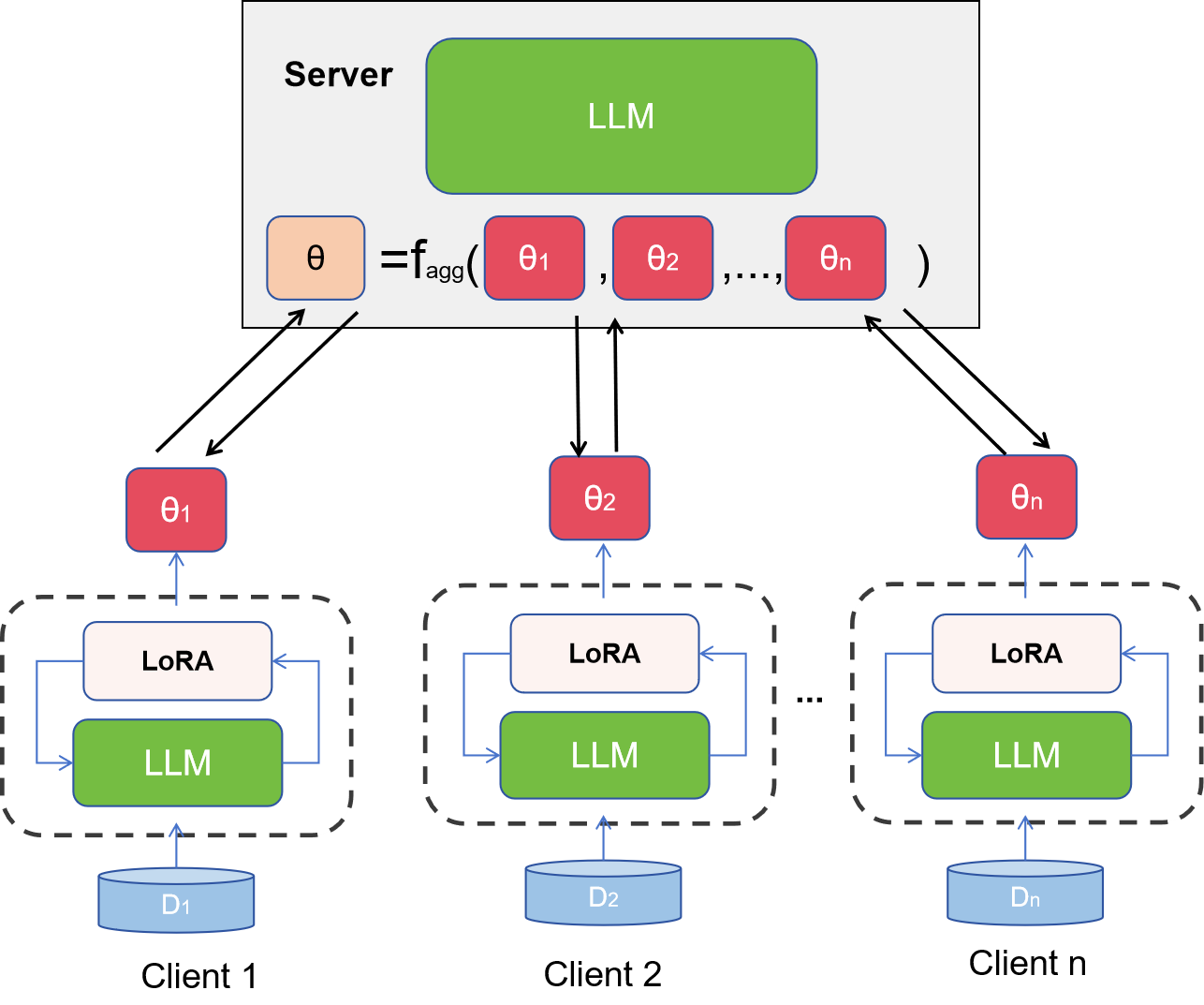}  
    \caption{This figure illustrates the training framework for federated large language models(federated LLMs). Due to limited computational resources on the client side, the client's LLM utilizes LoRA to train only a subset of its parameters.}  
    \label{fig:example}  
\end{figure}

Federated Large Language Models (federated LLMs) combine the advantages of FL and LLMs, enabling collaborative training of LLMs across distributed clients. This approach addresses challenges such as low data quality and high computational requirements typically associated with LLMs \cite{sani2024futurelargelanguagemodel}. Federated LLMs are especially beneficial in privacy-sensitive areas like healthcare and finance, where data cannot be centrally aggregated due to privacy restrictions.

The framework of federated LLM is shown in Figure \ref{fig:example}. Both the client and the server typically use LLMs with the same architecture. Due to limited computational resources on the client side, clients employ PEFT (Parameter Efficient Fine-Tuning) for training. Existing work \cite{fan2024fedcollm,fan2024fedmkt} has better addressed the issue of insufficient computational power on the client side by using LLMs on the server side and small language models(SLMs) with the same architecture on the client side. Knowledge is transferred from the server's LLM to the SLMs aggregated by the clients via knowledge distillation, which has proven effective. For simplicity, we assume that both the server and the client use LLMs with the same architecture and scale as \cite{10.1007/978-981-97-5569-1_17}. Like FL, federated LLMs undergo multiple rounds of training. In the federated LLMs training process, the server initially distributes its model parameters $\theta$ to clients. Each client then updates these parameters using its private dataset, subsequently returning the modified model parameters $\theta^k$ to the server, which combines the received updates to complete a training iteration.

However, as an emerging field, federated LLMs introduces new challenges, particularly those related to robustness. Existing research has yet to address these challenges. We will discuss the potential robustness challenges of federated LLMs in Section \ref{robustness problem}.

\subsection{Adversarial Training}

The adversarial training optimization problem in machine learning can be described as follows:
\begin{equation}
\label{equa:AT1}
\min_{\theta} \max_{\delta} \mathbb{E} \left[ \mathcal{L} \left( f_{\theta} \left( x + \delta \right), y \right) \right]
\end{equation}

Here, $\delta$ denotes the adversarial perturbation. The objective is to identify the $\delta$ that increases the model's loss via adversarial attacks. The model is subsequently trained on adversarial examples $x + \delta$, updating the parameters $\theta$ to enhance robustness.

Applying this adversarial training method to LLMs presents several challenges. First, the training data for LLMs is typically in discrete text space, which means that finding the optimal adversarial samples $x_{adv}$ requires traversing all possible tokens in the vocabulary to identify the adversarial sample that maximizes the model's loss. Although some methods \cite{mazeika2024harmbenchstandardizedevaluationframework} have improved the efficiency of this process, the computational cost remains high.

In the training and inference processes of LLMs, discrete prompts are first converted into discrete tokens. These tokens are then transformed into continuous embedding vectors through the model's embedding layer, placing them in what is known as the embedding space. Prior studies \cite{xhonneux2024efficientadversarialtrainingllms} have shown that creating adversarial examples within the embedding space effectively improves the robustness and rejection capabilities of large language models (LLMs). This capability refers to the model's ability to refuse to respond when encountering unsafe prompts. In addition, this approach significantly reduces training costs. 



\section{Robustness of Federated LLMs}\label{robustness problem}

Various challenges arise in different settings, as summarized in the table \ref{robustness-table}. Federated LLMs are susceptible to these challenges, which are inherited from both FL and LLM. The increased model scale amplifies the impact and complexity of these issues, while the synergy between FL and LLM also gives rise to novel robustness challenges. This section provides a comprehensive overview of the potential robustness challenges that federated LLMs may encounter.

\begin{table}[htbp]
\caption{Robustness in Different Settings}
\label{robustness-table}
\vskip 0.15in
\begin{center}
\begin{small}
\begin{sc}
\begin{tabular}{lccc}
\toprule
\textbf{Robust Challenges} & \textbf{FL} & \textbf{LLM} & \textbf{FedLLMs} \\
\midrule
Malicious Attacks         & $\surd$ & $\surd$ & \cellcolor{blue!20}{$\surd$} \\
Data Heterogeneity          & $\surd$ & & \cellcolor{blue!20}{$\surd$} \\
Errors and Outliers         & $\surd$ & $\surd$ & \cellcolor{blue!20}{$\surd$} \\
New Robust challenges                   & & & \cellcolor{blue!20}{$\surd$} \\
\bottomrule
\end{tabular}
\end{sc}
\end{small}
\end{center}
\vskip -0.1in
\end{table}

\subsection{Malicious Attacks}

Malicious attacks present a considerable risk to federated large language models, potentially undermining the global model's performance and stability. These threats can take multiple forms, including model poisoning, data poisoning, and adversarial attacks. In traditional FL, robust aggregation methods like the geometric median have been effective in mitigating malicious updates \cite{pillutla2022robust}, while adversarial training has shown promise in enhancing the robustness of LLMs \cite{mazeika2024harmbenchstandardizedevaluationframework,xhonneux2024efficientadversarialtrainingllms}. Although these methods have not been extensively validated in federated LLMs, their theoretical foundations suggest they could be valuable in addressing malicious attacks in this context. Future work should focus on experimentally validating these approaches in federated LLM settings.

\subsection{Data Heterogeneity}

Data heterogeneity is a core challenge in federated LLMs, arising from the diverse data distributions across clients. This issue is particularly pronounced in language models due to variations in language, domains, and corpora. Traditional federated learning has explored techniques like domain adaptation and multi-task learning to address data heterogeneity \cite{ghosh2019robustfederatedlearningheterogeneous}. Although these approaches have not been thoroughly explored in the context of federated LLMs, their ability to enhance generalization across diverse datasets suggests a promising avenue for future research.

\subsection{Errors and Outliers}

Federated LLMs are vulnerable to errors and outliers due to the varying quality of client data. Issues such as low-quality data, anomalous inputs, and faulty model updates can negatively impact the global model's performance. In conventional FL, various robust aggregation strategies and outlier detection techniques have been introduced to address these challenges. These methods offer valuable guidance for enhancing resilience against errors and outliers.

\subsection{New Robustness Challenges in Federated LLMs}

Federated LLMs introduce unique robustness challenges, such as cross-client update alignment, insufficient local computation resources, and the trade-off between communication compression and robustness. These challenges stem from the combination of FL and LLMs, particularly in high-dimensional parameter spaces. Techniques like gradient quantization have been explored in traditional federated learning to balance communication efficiency and robustness, but their application to federated LLMs requires further investigation. Future studies should aim to design adaptive approaches to tackle these challenges and maintain the stability of federated LLMs in decentralized settings.

These challenges demand innovative research into the robustness of federated LLMs. Future work should focus on developing targeted solutions in areas such as adversarial attack defense, resource-efficient aggregation strategies, and robustness enhancement techniques. In this article, we focus on the robustness of federated LLMs against malicious attacks and develops an algorithm aimed at improving the robustness of these models. 

\section{Threat Model}
In this study, the adversary evaluates the target model's robustness by crafting adversarial examples. By accessing the model's input space, the adversary introduces small perturbations \(\delta\) to generate adversarial samples in the form of \(x + \delta\), causing the model to produce incorrect predictions while ensuring the perturbations \(\|\delta\|_p \leq \epsilon\) remain imperceptible.

The adversary uses gradient-based attacks like FGSM or PGD to optimize perturbations, maximizing the loss function \(\mathcal{L}(f(x + \delta), y)\) with the dual objective of significantly degrading model performance on adversarial examples while maintaining high accuracy on clean data.

\section{Method}\label{sec3}
This section provides an in-depth introduction to the proposed FedEAT algorithm. We begin by discussing how to generate adversarial samples in the embedding space. We then introduce the robust aggregation algorithm employed in our approach. Finally, we outline the overall architecture of the FedEAT algorithm.

\subsection{Adversarial Training in the Embedding Space}

\begin{figure*}[htbp]
    \centering
    \includegraphics[width=0.8\textwidth]{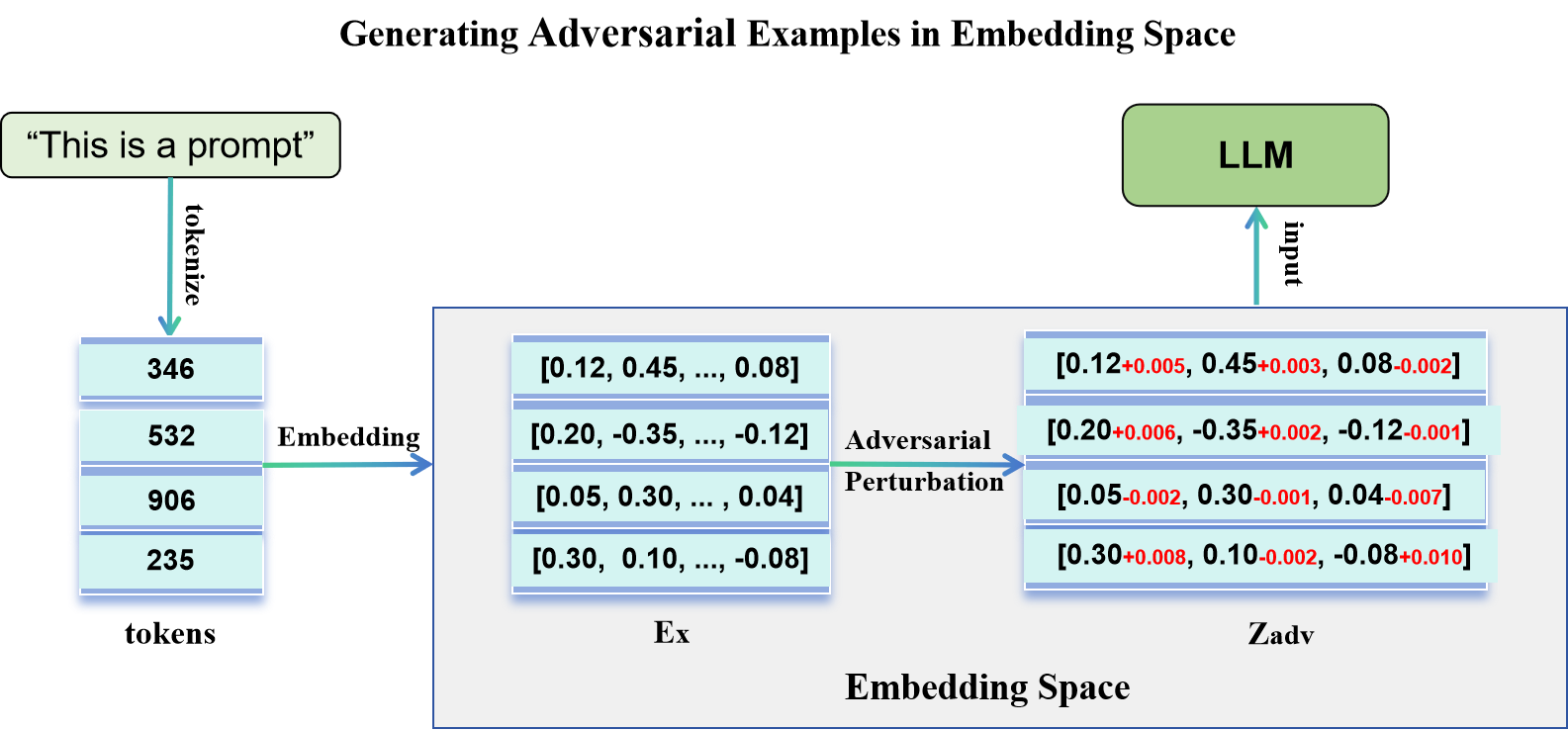}
    \caption{This figure depicts the process of generating adversarial samples in the embedding space. First, the input prompt is tokenized and then embedded into a vector space using the LLM's embedding layer. Next, a random perturbation is generated and iteratively updated using gradient ascent to maximize its adversarial effect. Finally, the perturbation is added to the original embedding vector to obtain the adversarial sample.}
    \label{fig:samples}
\end{figure*}

\cite{xhonneux2024efficientadversarialtrainingllms} introduces a harmful dataset containing harmful prompts, unsafe responses, and safe responses. By applying adversarial training in the embedding space using this dataset, the goal is to enable the LLM to identify more malicious prompts that induce unsafe responses, thereby generating safe responses to any harmful prompts. The results of \cite{xhonneux2024efficientadversarialtrainingllms} demonstrate that adversarial training in the embedding space can effectively enhance the LLM's robustness in rejecting such harmful inputs. Inspired by this work, we apply adversarial training in the embedding space to improve the robustness of federated LLMs. Unlike the aforementioned paper, our objective is to significantly enhance the robustness of federated LLMs without significantly compromising their performance. We achieve this by applying adversarial attacks in the embedding space on prompts from a fine-tuning dataset, thereby improving the LLM's robustness against adversarial attacks. The optimization problem is formulated as follows:

\begin{align}
\centering
\min_{\theta} & \left( \mathcal{L}(f(\mathbf{z}), y) - \lambda \cdot \max_{\delta \in \mathcal{S}} \mathcal{L}(f(\mathbf{z} + \delta), y) \right) \nonumber \\
\text{s.t.} \quad & \|\delta\|_p \leq \epsilon
\label{problem-goal}
\end{align}

The core of this optimization problem lies in achieving a balanced improvement of model robustness and performance through adversarial training. Specifically, the objective function comprises two components:
\begin{enumerate}
    \item The first term $\mathcal{L}(f(\mathbf{z}), y)$ represents the standard loss on the original input embedding vector $\mathbf{z}$, which ensures the model preserves its predictive accuracy on benign samples.
    
    \item The second term $-\lambda \cdot \max_{\delta \in \mathcal{S}} \mathcal{L}(f(\mathbf{z} + \delta), y)$ maximizes the loss induced by adversarial perturbation $\delta$, compelling the model to develop robustness against worst-case perturbations in the embedding space.
\end{enumerate}

The hyperparameter $\lambda$ governs the trade-off between robustness and standard performance, while the constraint $\|\delta\|_p \leq \epsilon$ maintains semantic consistency of adversarial examples by bounding the perturbation magnitude through the $L_p$-norm (typically with $p=2$ or $\infty$). This norm constraint prevents excessive distortion of input semantics that could otherwise degrade model functionality.


z+$\delta$ represents the adversarial example in the embedding space and the goal of adversarial examples is to find a perturbation $\delta$ in the embedding space that increases the loss of the model while keeping the perturbation within a certain norm to avoid significantly impacting performance. 

In each iteration of Projected Gradient Descent (PGD), the adversarial perturbation is updated by gradient ascent:

\begin{equation}
\label{equa:adv_sample}
\mathbf{z}^{t+1} = \mathbf{z}^t + \text{Proj}_{\mathcal{S}} \left(  \alpha \cdot \nabla \mathcal{L}(f(\mathbf{z}^t), y) \right) 
\end{equation}

At iteration \(t\), \(\mathbf{z}^t\) represents the embedding vector, updated using a step size \(\alpha\) and guided by the loss function gradient \(\nabla \mathcal{L}(f(\mathbf{z}^t), y)\). The \(\text{Proj}_{\mathcal{S}}\) operation constrains the adversarial perturbation within the predefined set \(\mathcal{S}\).

As shown in Figure \ref{fig:samples}, the process of generating adversarial samples in the embedding space involves several steps. First, the prompt is tokenized and then converted into an embedding vector $E_x$ via the model's embedding layer. Then, we introduce a random perturbation $\delta$ to $E_x$ and update it using gradient ascent to obtain the optimal perturbation $\delta$. This generates the adversarial samples $Z_{adv}$ in the embedding space. Finally, these samples are fed into the transformer layers for further processing.

\subsection{Geometric Median Aggregation of Client Parameters}\label{subsec-GM}
Inspired by \cite{pillutla2022robust}, we utilize a geometric median-based aggregation method to enhance the robustness of the global model. This method mitigates the impact of malicious or noisy client updates on the global model by implementing robust aggregation techniques, thereby enhancing the model's stability and resilience. This raises an important question in the research of federated LLM robustness:

\textbf{Does the robustness-enhancing aggregation method in federated learning remain effective in the federated LLMs setting?}

This question is crucial for research on the robustness of federated LLMs. 
In the FL setting, suppose there are \(n\) participating clients, after each round of training, we can obtain \( n \) sets of model parameters \( \{w_1, w_2, \dots, w_n\} \). The geometric median represents the optimal location within a dataset that achieves the minimal cumulative distance to all remaining data points. Therefore, the goal of geometric median aggregation is to find the global update that minimizes the total distance from all points to the aggregation point:

\begin{equation}
w_{\text{agg}} = \arg\min_w \sum_{k=1}^{n} \|w_k - w\|, 
\label{equa:geometric_median}
\end{equation}

where \( \|\cdot\| \) denotes the Euclidean norm. Due to the high computational complexity of directly solving for the geometric median, iterative methods are commonly used to approximate the optimal solution. The Weiszfeld algorithm is an efficient method for this purpose, with the update step given by:

\begin{equation}
\label{equa:GM_2}
w^{(t+1)} = \frac{\sum_{k=1}^{n} \frac{w_k}{\|w_k - w^{(t)}\|}}{\sum_{k=1}^{n} \frac{1}{\|w_k - w^{(t)}\|}}.
\end{equation}

This method iteratively updates \( w^{(t)} \) until convergence, providing an approximate solution to the geometric median.

In federated LLMs training, data distributions are often Non-IID, and some clients may upload anomalous or malicious updates. Simple averaging aggregation methods, such as FedAvg, have low robustness to noise and attacks. In contrast, the geometric median aggregation method is more robust to outliers and anomalies, significantly reducing the impact of malicious clients. This leads to improved accuracy and stability in the aggregation results, effectively addressing the challenges of malicious attacks and data heterogeneity mentioned in Section \ref{robustness problem}.

\subsection{Federated Embedding-space Adversarial Training: FedEAT}

Our algorithm introduces adversarial training in the embedding space into the training of federated LLMs to enhance the robustness of each local LLM. Additionally, to enhance both the effectiveness and resilience of the global model, we implement the geometric median-based aggregation approach.

\begin{algorithm}[tb]
   \caption{FedEAT (Client)}
   \label{alg:client}
\begin{algorithmic}[1]
   \STATE \textbf{Input:} Local dataset $D_k$, Global model parameters $\theta_{\text{global}}$
   \STATE \textbf{Parameters:} Learning rate $\eta$, Training rounds $E$
   \STATE $\theta_c \gets \theta_{\text{global}}$ \COMMENT{Initialize client model}
   \FOR{$e = 1$ \textbf{to} $E$}
        \FORALL{$(x, y) \in D_k$}
            \STATE $z \gets \text{Embed}(x)$ \COMMENT{Convert input to embedding}
            \STATE $e \gets  \text{Proj}_{\mathcal{S}} \left(  \alpha \cdot \nabla \mathcal{L}(f(\mathbf{z}^t), y) \right) $
            \STATE $z_{\text{adv}} \gets z + e$
            \STATE $L \gets \text{LossFunction}(\theta_c, z_{\text{adv}}, y)$
            \STATE $\theta_c \gets \theta_c - \eta \nabla_{\theta_c} L$ \COMMENT{Update model}
        \ENDFOR
   \ENDFOR
   \STATE \textbf{Return:} $\theta_c$ 
\end{algorithmic}
\end{algorithm}

\begin{algorithm}[tb]
\caption{FedEAT (Server)}
\label{alg:server}
\begin{algorithmic}[1]
\STATE \textbf{Input:} Starting global parameters $\mathbf{w}^{(0)}$, total rounds $T$, clients of each round $m$, convergence threshold $\epsilon$
\FOR{$t = 0, \ldots, T-1$}
\STATE Random select $m$ clients from $[n]$ to form set $S_t$
\FOR{each client $i \in S_t$ \textbf{in parallel}}
\STATE $\mathbf{w}_i^{(t+1)} \gets \text{FAT}(\mathbf{w}^{(t)})$ \COMMENT{Algorithm \ref{alg:client}}
\ENDFOR

   \REPEAT
       \STATE Update $\mathbf{w}^{(t+1)}$ using Weiszfeld algorithm as in Eq.(\ref{equa:GM_2})
   \UNTIL{convergence}
\ENDFOR
\STATE \textbf{Return:} $\mathbf{w}^{(T)}$
\end{algorithmic}
\end{algorithm}

On the client side (Algorithm \ref{alg:client}), each client initializes its model parameters $\theta_c$ with the global model parameters $\theta_{\text{global}}$. For each training epoch $e$, the client iterates over its local dataset $D$ and processes each sample $(x, y)$. 

First, the input data $x$ is transformed into an embedding vector $z$ through the embedding layer of the client LLM. Next, an adversarial perturbation $e$ is generated for the current client model $\theta_c$ and the embedding vector $z$ using multiple iterations of gradient ascent. Here, $\alpha$ represents the step size for updating the perturbation, and $\nabla$ represents the gradient of the embedding vector concerning the client LLM. To ensure that the perturbation does not excessively degrade the model's performance, it is restricted to a norm $S$ after each update.

The perturbation is subsequently applied to the embedding, resulting in the creation of an adversarial sample, expressed as $z_{\text{adv}} = z + e$, and the loss function $L$ is computed using this adversarial sample. Lastly, the model parameters $\theta_c$ are optimized using gradient descent with a learning rate $\eta$, improving the LLM's robustness to adversarial perturbations.

This process is repeated for $E$ epochs, after which the updated client model parameters $\theta_c$ are returned to the server.

On the server side (Algorithm \ref{alg:server}), the updated model parameters from all clients are aggregated. During each communication round, the server selects $m$ clients at random. Each selected client LLM performs local adversarial training in the embedding space in parallel, resulting in updated model parameters $\mathbf{w}_i^{(t+1)}$. 

To ensure robustness against potential malicious updates, the server employs the geometric median aggregation method. Specifically, the Weiszfeld algorithm (see Eq.(\ref{equa:GM_2})) is applied iteratively to compute the optimal parameters $\mathbf{w}^{(t+1)}$ until convergence. This process is repeated for $T$ rounds, after which the aggregated global model $\mathbf{w}^{(T)}$ is returned.

The FedEAT algorithm combines local adversarial training on clients with geometric median aggregation on the server. By introducing adversarial perturbations in the embedding space and employing robust aggregation, it enhances the robustness of federated large models while ensuring performance, thereby solving the optimization problem \ref{problem-goal}.

\section{Experiment}
\subsection{Experimental Setup}
\textbf{Dataset}: We utilize the \textit{vicgalle/alpaca-gpt4} dataset, based on which we generate adversarial samples for adversarial training of federated LLMs.

\textbf{Model}: We select the \textit{gemma-1.1-2b-it, PHI-3-MINI, MISTRAL-7B, and ZEPHYR-7B} for training and evaluating, as they support direct training through embedding vectors.

\textbf{Evaluation Metrics}: We employ a comprehensive evaluation framework to assess the utility and robustness of federated LLMs. Specifically, we utilize the benign and adversarial datasets from \cite{huang2024trustllm} to evaluate the models' performance on four tasks, including SST2, QQP, MNLI, and QNLI. Our evaluation metrics include classic measures of utility and robustness, and robustness describes a model's ability to maintain performance despite noise or perturbations.

The Attack Success Rate (ASR) measures an LLM's robustness by calculating the proportion of benignly classified samples that become misclassified under adversarial conditions, expressed as $\text{ASR} = \frac{A_m}{B_c}$, where $B_c$ is the number of correctly classified samples and $A_m$ is the count of those samples misclassified under adversarial perturbations.

A lower ASR indicates that the model can maintain its performance under adversarial perturbations, i.e., the model is more robust and resistant to attacks, and is less likely to change its correct judgment due to small changes or malicious perturbations in the data. Conversely, a higher ASR means that the model is more prone to errors when faced with perturbations, indicating lower robustness.

Evaluation of robustness methods and examples of benign and adversarial datasets are provided in the appendix \ref{Benign&Adversarial_Datasets}.

\subsection{Results of Robustness and Utility}

\begin{table}[htbp]
\caption{Robustness of Different Methods on Various Tasks}
\label{robustness-table}
\vskip 0.15in
\begin{center}
\begin{small}
\begin{sc}
\begin{tabular}{lcccc}
\toprule
\textbf{Task} & \textbf{Method} & \textbf{PHI-3-MINI} & \textbf{ZEPHYR-7B} \\
\midrule
\rowcolor[gray]{0.9}  
SST2 & FedAVG & 0.436 & \textbf{0.213} \\
     & FedEAT & \textbf{0.436} & 0.226 \\
\rowcolor[gray]{0.9}  
QQP  & FedAVG & 0.170 & 0.150 \\
     & FedEAT & \textbf{0.170} & \textbf{0.060} \\
\rowcolor[gray]{0.9}  
MNLI & FedAVG & 0.267 & 0.145 \\
     & FedEAT & \textbf{0.256} & \textbf{0.121} \\
\rowcolor[gray]{0.9}  
QNLI & FedAVG & 0.162 & 0.196 \\
     & FedEAT & \textbf{0.146} & \textbf{0.168} \\
\midrule
\rowcolor[gray]{0.9}  
Average & FedAVG & 0.259 & 0.176 \\
        & FedEAT & \textbf{0.252} & \textbf{0.144} \\
\bottomrule
\end{tabular}
\end{sc}
\end{small}
\end{center}
\vskip -0.1in
\end{table}

\begin{figure}[htbp]
    \centering
    \includegraphics[width=0.48\textwidth]{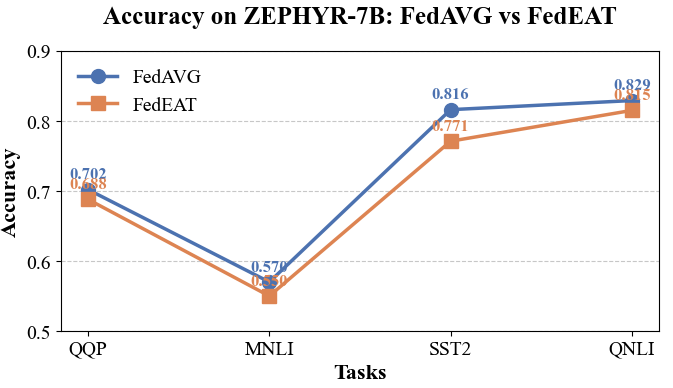}
    \caption{The figure presents a comparison of accuracy on a benign dataset between models trained using the FedAVG and FedEAT methods within a federated LLMs framework based on the ZEPHYR-7B. Accuracy is used as a metric to assess the utility of these methods.}
    \label{fig:utility}
\end{figure}

As illustrated in Figure \ref{fig:utility}, the test accuracy on the benign dataset indicates that the FedEAT method incurs only minimal utility loss compared to FedAVG. In some cases, it even enhances the model's generalization ability, thereby improving utility. This suggests that adversarial training in the embedding space not only boosts robustness but also helps the model learn more generalizable features. The complete results can be found in Appendix 1. Additionally, the results on the adversarial dataset (Table \ref{robustness-table}) demonstrate that the LLMs trained with the FedEAT method exhibit significantly better robustness compared to those trained with the FedAVG method. The complete results are available in Appendix 1. 

This improvement can be attributed to two key components of FedEAT: (1) adversarial training in the embedding space, which enhances the model's resistance to adversarial perturbations, and (2) the geometric median aggregation method, which effectively mitigates the impact of noisy or malicious updates from clients. In summary, these findings underscore FedEAT's capability to maintain a balance between utility and robustness, positioning it as a promising strategy for deploying federated LLMs in real-world scenarios.

\begin{figure*}[htbp]
    \centering
    \includegraphics[width=0.48\textwidth]{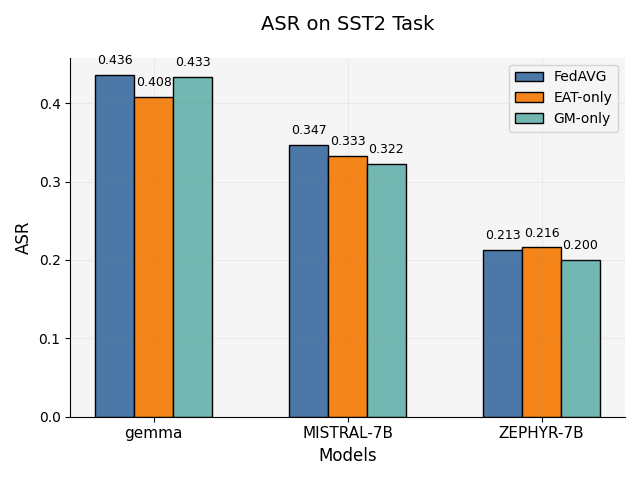}  
    \hfill 
    \includegraphics[width=0.48\textwidth]{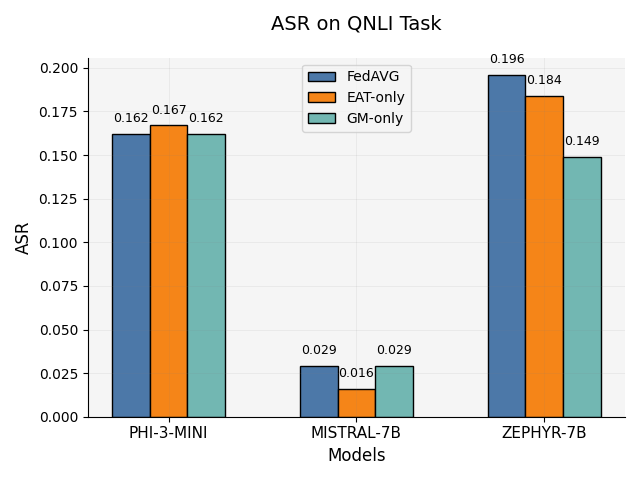}  
    \caption{Ablation Study on Robustness Improvement: The Impact of Embedding-space Adversarial Training and Geometric Median Aggregation in SST2 and QNLI Tasks. The ASR for both EAT-only and GM-only is lower than that of FedAVG, indicating that both embedding-space adversarial training and geometric median aggregation methods are effective in enhancing the robustness of federated LLMs.}
    \label{fig:combined}
\end{figure*}

\subsection{Ablation Study}

To validate the effectiveness of the two proposed methods in our algorithm for enhancing the robustness of LLMs, we trained LLMs using only the embedding space adversarial training (EAT-only) method and only the geometric median aggregation (GM-only) method, and compared them to models trained using the standard FedAVG method.

As shown in Figure \ref{fig:combined}, for multiple tasks across the two models, the ASR of LLMs trained with EAT-only and GM-only methods is lower than that of LLMs trained with FedAVG in most tasks. This indicates that both EAT-only and GM-only models exhibit superior robustness compared to the FedAVG model. This confirms the issue we raised in Section \ref{subsec-GM}, demonstrating that robust aggregation algorithms in federated learning can still effectively enhance the robustness of federated LLMs.

\section{Conclusion}\label{sec4}
Federated LLMs hold significant promise for the future. This paper explores the potential robustness challenges faced by federated LLMs and proposes the FedEAT algorithm to enhance their robustness. Experimental results demonstrate that FedEAT improves model robustness with only a minor performance trade-off, and the ablation studies validate our algorithm's effectiveness.


\nocite{langley00}

\bibliography{example_paper}
\bibliographystyle{icml2025}

\newpage
\appendix
\onecolumn
\section{Appendix}
\subsection{Limitation}
Due to limitations in some models' ability to support training and inference with embedding vectors, our method has only been tested on models compatible with this approach. In the future, we plan to develop new techniques to enhance the universality of our method. Additionally, our current experiments were conducted on the \textit{vicgalle/alpaca-gpt4} dataset, and we intend to extend our evaluations to include more models and datasets.

Currently, our robustness evaluation is based on adding noise to the dataset and measuring the model's accuracy. However, we plan to employ a variety of evaluation methods to conduct a more comprehensive robustness assessment of the trained LLMs. This will help verify the improvements in model robustness brought by our algorithm and provide a more thorough understanding of its effectiveness in real-world applications.

Since LLMs with different architectures exhibit varying sensitivities to adversarial perturbations, and our experiments applied the same perturbation intensity to all LLMs, the results may not fully reflect the algorithm's effectiveness. In subsequent work, we will conduct additional experiments to identify the optimal perturbation intensity for different LLMs.

\subsection{Complete results of different methods applied to various tasks and models.}\label{subsec-results}

\begin{table*}[htbp]
\caption{Robustness of Different Methods on Various Tasks}
\label{robustness-table-all}
\vskip 0.15in
\begin{center}
\begin{small}
\begin{sc}
\begin{tabular}{lccccc}
\toprule
\textbf{Task} & \textbf{Method} & \textbf{gemma} & \textbf{PHI-3-MINI} & \textbf{MISTRAL-7B} & \textbf{ZEPHYR-7B} \\
\midrule
\rowcolor[gray]{0.9} 
SST2 & FedAVG & 0.436 & 0.436 & 0.347 & \textbf{0.213} \\
     & FedEAT & \textbf{0.425} & \textbf{0.436} & \textbf{0.304} & 0.226 \\
\rowcolor[gray]{0.9} 
QQP  & FedAVG & 0    & 0.170 & \textbf{0.034} & 0.150 \\
     & FedEAT & 0    & \textbf{0.170} & 0.045 & \textbf{0.060} \\
\rowcolor[gray]{0.9} 
MNLI & FedAVG & 0    & 0.267 & \textbf{0.190} & 0.145 \\
     & FedEAT & 0    & \textbf{0.256} & 0.300 & \textbf{0.121} \\
\rowcolor[gray]{0.9} 
QNLI & FedAVG & 0    & 0.162 & 0.029 & 0.196 \\
     & FedEAT & 0    & \textbf{0.146} & \textbf{0.015} & \textbf{0.168} \\
\midrule
\rowcolor[gray]{0.9} 
Average & FedAVG & 0.109 & 0.259 & 0.150 & 0.176 \\
        & FedEAT & \textbf{0.106} & \textbf{0.252} & \textbf{0.166} & \textbf{0.144} \\
\bottomrule
\end{tabular}
\end{sc}
\end{small}
\end{center}
\vskip -0.1in
\end{table*}

The table \ref{robustness-table-all} presents the robustness experimental results of different methods (FedAVG and FedEAT) across multiple tasks (SST2, QQP, MNLI, QNLI) and various models (GEMMA, PHI-3-MINI, MISTRAL-7B, ZEPHYR-7B). The values in the table represent the Attack Success Rate (ASR), where lower values indicate better model robustness.

From the results, it is evident that FedEAT demonstrates superior robustness compared to FedAVG in most tasks and models. For instance, in the SST2 task, FedEAT achieves lower ASR than FedAVG on the GEMMA, PHI-3-MINI, and MISTRAL-7B models. In the QQP task, FedEAT also shows lower ASR than FedAVG on the PHI-3-MINI and ZEPHYR-7B models. Similarly, in the MNLI and QNLI tasks, FedEAT outperforms FedAVG across multiple models.

Particularly noteworthy is the fact that the ASR for the GEMMA model is 0 in the QQP, MNLI, and QNLI tasks. This is because the GEMMA model consistently outputs "yes" for yes-or-no decision problems, regardless of the input, rendering the results non-representative. We plan to address this issue in future work to improve the accuracy of robustness evaluation for the GEMMA model on these tasks.

Overall, FedEAT demonstrates excellent performance in enhancing model robustness, with its average ASR across multiple tasks and models being lower than that of FedAVG. This indicates that FedEAT has significant advantages in improving the robustness of federated LLMs.

\begin{table*}[htbp]
\caption{Accuracy of Different Methods on Various Tasks}
\label{utility-table-all}
\vskip 0.15in
\begin{center}
\begin{small}
\begin{sc}
\begin{tabular}{lccccc}
\toprule
\textbf{Task} & \textbf{Method} & \textbf{gemma} & \textbf{PHI-3-MINI} & \textbf{MISTRAL-7B} & \textbf{ZEPHYR-7B} \\
\midrule
\rowcolor[gray]{0.9} 
SST2 & FedAVG & 0.840 & 0.840 & \textbf{0.901} & \textbf{0.816} \\
     & FedEAT & \textbf{0.916} & \textbf{0.840} & 0.954 & 0.771 \\
\rowcolor[gray]{0.9} 
QQP  & FedAVG & 0.423 & 0.746 & 0.547 & \textbf{0.702} \\
     & FedEAT & 0.423 & 0.746 & \textbf{0.595} & 0.688 \\
\rowcolor[gray]{0.9} 
MNLI & FedAVG & 0.264 & 0.711 & 0.521 & \textbf{0.570} \\
     & FedEAT & 0.264 & 0.711 & \textbf{0.744} & 0.550 \\
\rowcolor[gray]{0.9} 
QNLI & FedAVG & 0.526 & \textbf{0.789} & \textbf{0.531} & \textbf{0.829} \\
     & FedEAT & 0.526 & 0.721 & 0.512 & 0.815 \\
\midrule
\rowcolor[gray]{0.9} 
Average & FedAVG & 0.513 & \textbf{0.772} & 0.625 & \textbf{0.729} \\
     & FedEAT & \textbf{0.532} & 0.755 & \textbf{0.701} & 0.706 \\
\bottomrule
\end{tabular}
\end{sc}
\end{small}
\end{center}
\vskip -0.1in
\end{table*}

The table \ref{utility-table-all} presents the test accuracy of different models on benign datasets across multiple tasks. The results indicate that the FedEAT method incurs only minimal utility loss compared to FedAVG. In some cases, if the adversarial perturbations are relatively small for a specific LLM architecture, FedEAT can even enhance the model's generalization ability, thereby improving utility. This phenomenon occurs because different LLM architectures exhibit varying sensitivities to adversarial perturbations. Since we used a fixed perturbation intensity in our experiments, some models may benefit from smaller perturbations, leading to improved generalization, while others may experience performance degradation due to excessive perturbations. In future work, we plan to conduct additional experiments to identify the optimal perturbation intensity for different LLM architectures, further enhancing the overall performance of the models.

\begin{table*}[htbp]
\caption{Robustness of EAT and GM Aggregation Methods}
\label{Ablation-table-all}
\vskip 0.15in
\begin{center}
\begin{small}
\begin{sc}
\begin{tabular}{lccccc}
\toprule
\textbf{Task} & \textbf{Method} & \textbf{gemma} & \textbf{PHI-3-MINI} & \textbf{MISTRAL-7B} & \textbf{ZEPHYR-7B} \\
\midrule
\rowcolor[gray]{0.9} 
SST2 & FedAVG & 0.436 & 0.436 & 0.347 & 0.213 \\
     & EAT-only & \textbf{0.408} & \textbf{0.436} & \textbf{0.333} & 0.216 \\
     & GM-only & \textbf{0.433} & \textbf{0.436} & \textbf{0.322} & \textbf{0.200} \\
\rowcolor[gray]{0.9} 
QQP  & FedAVG & 0    & 0.170 & 0.034 & 0.150 \\
     & EAT-only & 0    & \textbf{0.170} & 0.083 & \textbf{0.086} \\
     & GM-only & 0    & \textbf{0.170} & 0.071 & 0.182 \\
\rowcolor[gray]{0.9} 
MNLI & FedAVG & 0    & 0.267 & 0.190 & 0.145 \\
     & EAT-only & 0    & \textbf{0.253} & 0.264 & \textbf{0.129} \\
     & GM-only & 0    & \textbf{0.256} & 0.257 & \textbf{0.145} \\
\rowcolor[gray]{0.9} 
QNLI & FedAVG & 0    & 0.162 & 0.029 & 0.196 \\
     & EAT-only & 0    & 0.167 & \textbf{0.016} & \textbf{0.184} \\
     & GM-only & 0    & \textbf{0.162} & \textbf{0.029} & \textbf{0.149} \\
\midrule
\rowcolor[gray]{0.9} 
Average & FedAVG & 0.109 & 0.259 & 0.150 & 0.176 \\
     & EAT-only & \textbf{0.102} & \textbf{0.256} & 0.174 & \textbf{0.164} \\
     & GM-only & \textbf{0.108} & \textbf{0.256} & 0.170 & \textbf{0.169} \\
\bottomrule
\end{tabular}
\end{sc}
\end{small}
\end{center}
\vskip -0.1in
\end{table*}

The table \ref{Ablation-table-all} presents the complete results of ablation experiments validating the effectiveness of Embedding-space Adversarial Training (EAT) and Geometric Median Aggregation (GM). Although the Attack Success Rate (ASR) for the GEMMA model is 0 in the QQP, MNLI, and QNLI tasks (as explained earlier), the results still clearly demonstrate that both the EAT-only (Embedding-space Adversarial Training only) and GM-only (Geometric Median Aggregation only) methods significantly enhance the robustness of federated large models. These methods achieve lower ASR compared to FedAVG in most tasks and models, further validating their effectiveness in improving model robustness.

\subsection{Examples of Benign and Adversarial Datasets for Robustness Evaluation}\label{Benign&Adversarial_Datasets}

\begin{table*}[htbp]
\caption{Comparison of Benign (Original) and Adversarial (Modified) Dataset Examples}
\label{tab:dataset-examples}
\vskip 0.15in
\begin{center}
\begin{small}
\begin{sc}
\begin{tabular}{lllll}
\toprule
\textbf{Index} & \textbf{Task} & \textbf{Type} & \textbf{Prompt (Question/Sentence)} & \textbf{Label} \\
\midrule
1 & QQP & \textbf{Original} & 
\begin{minipage}{0.4\textwidth}
\lowercase{question1: can eating only fruit for dinner lead to weight loss ? \\
question2: does eating dinner earlier in the evening help with weight loss ?}
\end{minipage} & 0 (No) \\
\addlinespace 
\rowcolor[gray]{0.9} 
 & & \textbf{Modified} & 
\begin{minipage}{0.4\textwidth}
\lowercase{question1: can eating precisely advance for dinner lead to weight loss ? \\
question2: does eating dinner earlier in the evening help with weight loss ?}
\end{minipage} & 0 (No) \\
\midrule
187 & SST2 & \textbf{Original} & 
\begin{minipage}{0.4\textwidth}
\lowercase{sentence: a subtle and well-crafted ( for the most part ) chiller .}
\end{minipage} & 1 (Positive) \\
\addlinespace 
\rowcolor[gray]{0.9} 
 & & \textbf{Modified} & 
\begin{minipage}{0.4\textwidth}
\lowercase{sentence: a subtle and well-crafted ( for the most part ) chiller and false is not true .}
\end{minipage} & 1 (Positive) \\
\midrule
300 & QNLI & \textbf{Original} & 
\begin{minipage}{0.4\textwidth}
\lowercase{question: what kind of needlework was used in the creation of the bayeux tapestry ? \\
answer: it was commissioned by odo , the bishop of bayeux and first earl of kent , employing natives from kent who were learned in the nordic traditions imported in the previous half century by the danish vikings .}
\end{minipage} & 1 (Yes) \\
\addlinespace 
\rowcolor[gray]{0.9} 
 & & \textbf{Modified} & 
\begin{minipage}{0.4\textwidth}
\lowercase{question: what kind of needlework was used in the creation of the bayeux tapestry ? \\
answer: maritime was sewing plates odo , the bishop of bayeux and first earl of kent , employing natives from kent who were learned in the nordic traditions imported in the previous half century by the danish vikings .}
\end{minipage} & 1 (Yes) \\
\bottomrule
\end{tabular}
\end{sc}
\end{small}
\end{center}
\vskip -0.1in
\end{table*}

The table \ref{tab:dataset-examples} presents a comparison of examples from benign (Original) and adversarial (Modified) datasets, aiming to evaluate the robustness and utility of models across different tasks and datasets. It includes examples from three tasks: QQP (semantic equivalence judgment), SST2 (sentiment analysis), and QNLI (question-answering natural language inference). For each task, the table provides a comparison between benign and adversarial data, clearly illustrating how adversarial attacks are generated through subtle semantic changes or the addition of disruptive information.

In the QQP task, the original question, "\textit{Can eating only fruit for dinner lead to weight loss?}" is modified to "\textit{Can eating precisely advance for dinner lead to weight loss?}" by replacing key phrases to test the model's sensitivity to semantic variations. In the SST2 task, the original sentence, "\textit{a subtle and well-crafted chiller,}" is augmented with the disruptive phrase "\textit{and false is not true}" to evaluate the model's tolerance to noise. In the QNLI task, the original answer, "\textit{it was commissioned by Odo,}" is altered to "\textit{Maritime was sewing plates Odo,}" introducing semantic changes to test the model's robustness.

The Modified rows are highlighted with a light gray background, making it easy to compare them with the Original rows. Each example is annotated with a label (e.g., 0 or 1), indicating the correct answer to the question or the sentiment of the sentence. These labels remain consistent between the original and adversarial data to ensure a fair evaluation. Through this comparison, the table visually demonstrates how adversarial datasets are generated through subtle modifications, providing valuable insights for assessing model robustness.


\end{document}